\documentclass[10pt,twocolumn,letterpaper]{article}

\usepackage{cvpr}              % To produce the CAMERA-READY version
\usepackage{amsfonts}
\usepackage{amsmath}
\usepackage{multirow}
\usepackage{amssymb}
\usepackage{pifont}
\usepackage{xcolor}
\usepackage{colortbl}
\usepackage{makecell}

\newcommand{\cmark}{\ding{51}}
\newcommand{\xmark}{\ding{55}}
\newcommand\blfootnote[1]{%
  \begingroup
  \renewcommand\thefootnote{}\footnote{#1}%
  \addtocounter{footnote}{-1}%
  \endgroup
}
\definecolor{tabfirst}{rgb}{1, 0.7, 0.7}
\definecolor{tabsecond}{rgb}{1, 0.85, 0.7}
\definecolor{tabthird}{rgb}{1, 1, 0.7}

\definecolor{cvprblue}{rgb}{0.21,0.49,0.74}
\usepackage[pagebackref,breaklinks,colorlinks,allcolors=cvprblue]{hyperref}

\def\paperID{} % *** Enter the Paper ID here
\def\confName{CVPR}
\def\confYear{2026}

\title{MoVieDrive: Urban Scene Synthesis with \\Multi-Modal Multi-View Video Diffusion Transformer}

\author{Guile Wu\textsuperscript{1}, David Huang\textsuperscript{1,2}{\footnotemark[1]}, Dongfeng Bai\textsuperscript{1}, and Bingbing Liu\textsuperscript{1}\\
\textsuperscript{1}Huawei Noah's Ark Lab \quad \textsuperscript{2}University of Toronto \\
{\tt\small guile.wu@outlook.com, dawae.huang@mail.utoronto.ca, \{baidongfeng, liu.bingbing\}@huawei.com}
}

\begin{document}
% \maketitle

\twocolumn[{%
\renewcommand\twocolumn[1][]{#1}%
\maketitle
% \vspace{-1cm}
\begin{center}
    \centering
    \captionsetup{type=figure}
    \includegraphics[height=7cm]{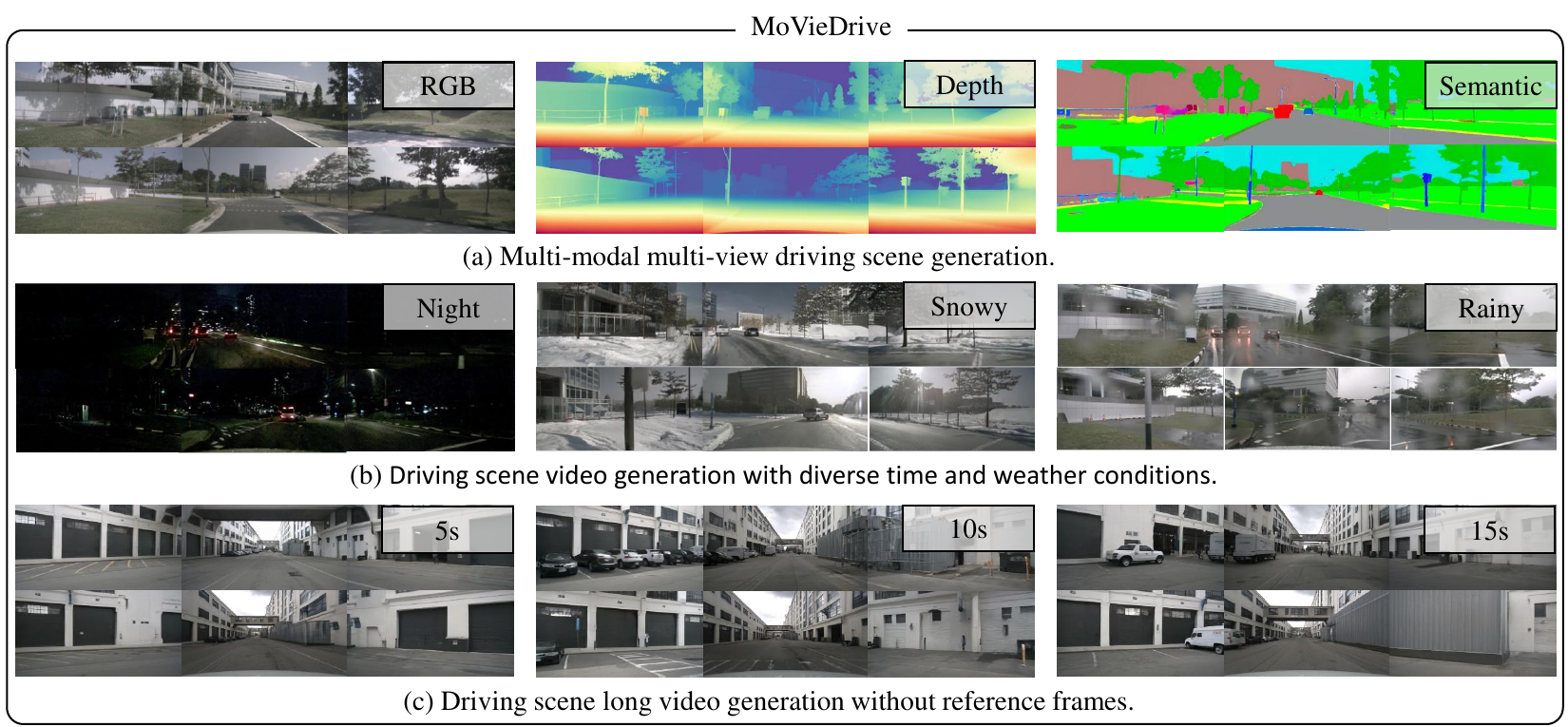}
    \vspace{-0.3cm}
    \captionof{figure}{
        An illustration of the proposed MoVieDrive approach to urban scene video generation in autonomous driving.
        Our approach can be used to generate multi-modal multi-view driving scene videos, to synthesize diverse driving scenes under diverse time and weather conditions, and to generate long videos without reference frames.
        }
    \label{fig_illustration}
\end{center}%
}]

\blfootnote{\textsuperscript{*}David Huang contributed to this work during an internship at Huawei Canada.}

\begin{abstract}
    Urban scene synthesis with video generation models has recently shown great potential for autonomous driving.
    Existing video generation approaches to autonomous driving primarily focus on RGB video generation and lack the ability to support multi-modal video generation.
    However, multi-modal data, such as depth maps and semantic maps, are crucial for holistic urban scene understanding in autonomous driving.
    Although it is feasible to use multiple models to generate different modalities, this increases the difficulty of model deployment and does not leverage complementary cues for multi-modal data generation.
    To address this problem, in this work, we propose a novel multi-modal multi-view video generation approach to autonomous driving.
    Specifically, we construct a unified diffusion transformer model composed of modal-shared components and modal-specific components.
    Then, we leverage diverse conditioning inputs to encode controllable scene structure and content cues into the multi-modal multi-view unified diffusion model.
    In this way, our approach is capable of generating multi-modal multi-view driving scene videos in a unified framework.
    Our thorough experiments on real-world autonomous driving dataset show that our approach achieves compelling video generation quality and controllability compared with state-of-the-art methods, while supporting multi-modal multi-view data generation.    
\end{abstract}    
\section{Introduction}
\label{sec:introduction}
Urban scene video generation for autonomous driving has advanced rapidly in recent years.
It can be used to generate controllable driving scenes, especially to synthesize long-tail scenarios that cannot be easily collected in the real world.
This facilitates performance improvement and reliability evaluation in autonomous driving.

Contemporary video generation methods, such as SVD~\cite{blattmann2023stable} and CogVideoX~\cite{yang2025cogvideox}, have shown promising performance for generating high-quality videos.
However, they cannot be directly used for autonomous driving because multi-view driving scene generation requires multi-view spatiotemporal consistency and high controllability.
To address this problem, some recent studies have explored diffusion models~\cite{ho2020denoising,rombach2022high} for controllable multi-view urban scene generation and shown promising performance~\cite{wang2024drivedreamer,wen2024panacea,gao2025magicdrive-v2,ni2025maskgwm}. 
However, these methods mostly focus on single-modal RGB video generation and lack the ability to support multi-modal video generation.
As a perception-intensive task, autonomous driving has an inherent need for multi-modal data, such as depth maps and semantic maps, which can facilitate holistic urban scene understanding towards more efficient self-driving.
A common approach to solving this problem is using multiple models to generate different modalities, but this often increases the difficulty of model deployment and does not fully exploit complementary cues for multi-modal data generation.

\begin{figure*}[t]
\centering
\includegraphics[width=0.92\textwidth]{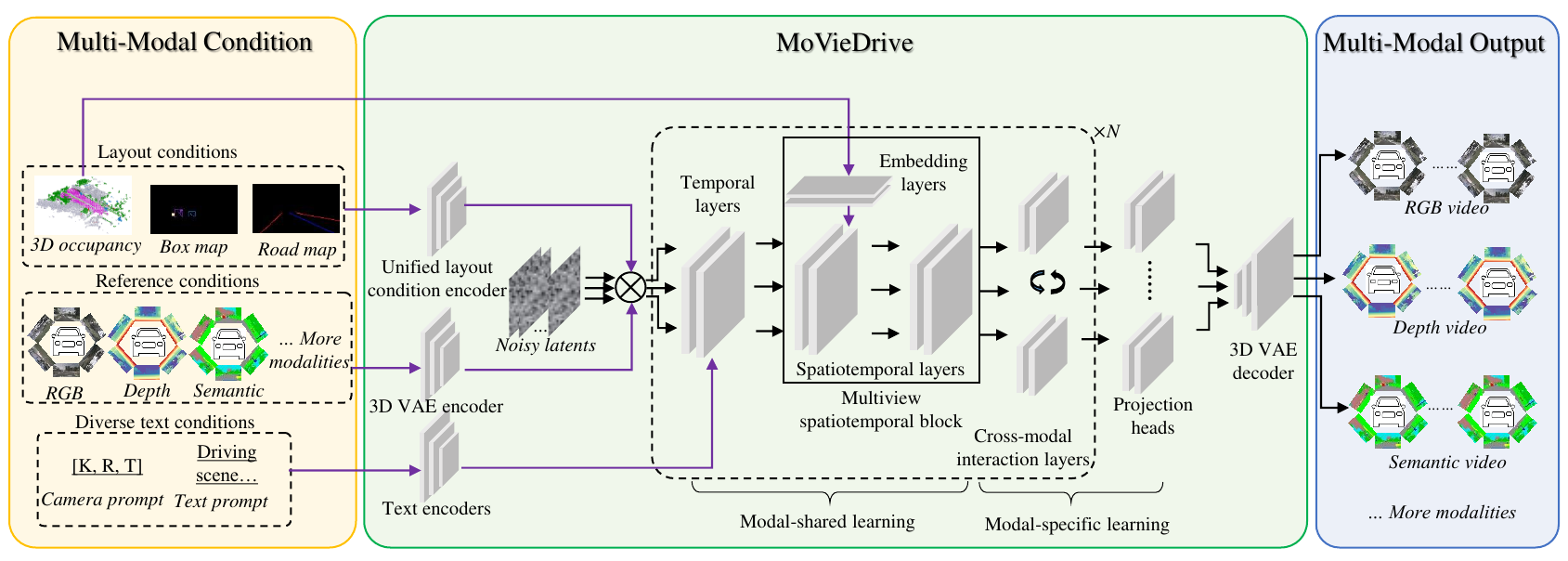}
\vspace{-0.3cm}
\caption{An overview of the proposed MoVieDrive approach.
Our approach employs diverse conditioning inputs and a multi-modal multi-view diffusion transformer model to facilitate urban scene understanding in autonomous driving.}
\label{fig_framework}
\end{figure*}

In this work, we propose a novel multi-\textbf{\emph{MO}}dal multi-\textbf{\emph{VIE}}w video diffusion approach, dubbed \textbf{\emph{MoVieDrive}}, to autonomous driving.
Instead of using multiple models to generate different modalities, our approach proposes to employ a unified model for multi-modal multi-view driving scene generation.
An illustration of our approach to urban scene video generation in autonomous driving is depicted in Fig.~\ref{fig_illustration}.
Specifically, we employ diverse conditioning inputs, including text conditions, context reference conditions, and layout conditions, to guide controllable scene generation.
Among them, text conditions guide holistic scene generation, context reference conditions are optional and only used for future scene prediction, and layout conditions capture fine-grained scene cues.
Next, we decompose multi-modal multi-view scene generation into modal-shared learning and modal-specific learning, and devise a unified diffusion transformer model with modal-shared layers and modal-specific layers.
This unified model alleviates the need for multiple models and improves model scalability.
We then encode diverse conditions into the unified diffusion model for multi-modal multi-view scene generation.
This helps to leverage complementary cues from multi-modal data for richer scene understanding to facilitate multi-modal multi-view urban scene generation in a unified framework.
Experimental results on real-world autonomous driving dataset demonstrate that our approach can generate multi-modal multi-view urban scene videos with high fidelity and controllability, surpassing state-of-the-art methods.
In summary, our \emph{\textbf{contributions}} are:

\begin{itemize}
    \item {We propose a novel framework that exploits diverse conditioning inputs and a diffusion transformer model for multi-modal multi-view autonomous driving scene generation.
    This fills a gap left by existing methods.}
    \item {We introduce a novel unified multi-modal multi-view video diffusion transformer model composed of modal-shared components and modal-specific components model for urban scene video generation.}
    \item {We provide thorough experiments on real-world autonomous driving dataset and demonstrate the superiority of our approach over state-of-the-art methods.}
\end{itemize}

\section{Related Work}
\label{sec:related_work}
\paragraph{Video Generation.}
As a long-standing research topic in computer vision, video generation has been well developed and various types of methods have been explored, \eg, diffusion models~\cite{ho2022video}, Variational Autoencoders (VAEs)~\cite{babaeizadeh2018stochastic}, generative adversarial networks~\cite{tulyakov2018mocogan}, autoregressive models~\cite{weissenborn2019scaling}, etc.
Among them, diffusion-based methods~\cite{blattmann2023stable,yang2025cogvideox,ho2022video,hong2023cogvideo} have recently become predominant due to their superiority in controllability and fidelity.
However, these methods cannot be directly applied to urban scene generation for autonomous driving, which requires multi-view spatiotemporal consistency and high controllability.
To adapt them to complex controllable driving scene generation, substantial modifications are needed~\cite{wang2024drivedreamer,gao2024magicdrive}.
Our approach shares the merits of diffusion-based video generation and devises a new multi-modal multi-view diffusion transformer model for autonomous driving.

\vspace{-0.5cm}
\paragraph{Urban Scene Synthesis.}
The rapid advancement in urban scene and asset generation has significantly facilitated data synthesis and closed-loop evaluation in autonomous driving~\cite{yang2023unisim,zhou2024hugs,gao2024magicdrive,liu2024vqa}.
There are two primary research directions in urban scene synthesis for autonomous driving.
One line of work explores rendering methods, such as neural radiance fields and 3D Gaussian splatting, for urban scene synthesis~\cite{yang2023unisim,zhou2024hugs,ren2024unigaussian,wu2025armgs}.
Despite high-fidelity scene synthesis, this type of method has poor controllability and diversity.
Another line of work explores video generation, such as diffusion models~\cite{blattmann2023stable,yang2025cogvideox}, for urban scene generation~\cite{wen2024panacea,wang2024drivedreamer}.
Among them, some studies focus on single-view generation~\cite{gao2024vista,wang2024drivingdojo}, while others investigate multi-view generation~\cite{gao2024magicdrive,wang2024driving,ni2025maskgwm}.
Our work focuses on the more challenging multi-view setting.
Contemporary methods, such as DriveDreamer~\cite{wang2024drivedreamer}, MagicDrive~\cite{gao2024magicdrive}, MagicDrive-V2~\cite{gao2025magicdrive-v2}, MaskGWM~\cite{ni2025maskgwm}, etc., widely use diffusion models for multi-view urban scene generation.
Although these methods have shown promising results, they only focus on RGB video generation and lack the ability to generate multi-modal data for holistic scene understanding.
UniScene~\cite{li2025uniscene} proposes using multiple models to generate RGB videos and LiDAR point clouds, but still fails to construct a unified model for multi-view multi-modal urban scene generation.
Unlike existing works, our approach proposes to generate multi-modal multi-view driving scene videos in a unified framework, which fills a gap left by existing methods.

\vspace{-0.5cm}
\paragraph{Multi-Modal Synthesis.}
There have been several diffusion-based multi-modal synthesis methods proposed in recent years~\cite{liu2024hyperhuman,zhai2024idol,stan2023ldm3d,xi2025omnivdiff}.
However, none of them are designed for urban scene generation and modifying them for complex controllable multi-modal multi-view scene generation is not trivial.
This often requires modifying conditioning input encoders, diffusion transformer layers, and training processes.
Note that investigating how to modify these approaches to autonomous driving is beyond the scope of this work.
Our approach differs from these works in that we propose a novel framework that uses diverse conditioning inputs to encode controllable scene structure and content cues into a diffusion transformer model for multi-modal multi-view driving scene generation.

\section{Methodology}
\label{sec:methodology}

\subsection{Approach Overview}

\paragraph{Problem Statement.}
This work focuses on multi-modal multi-view urban scene video generation for autonomous driving.
Specifically, our goal is to learn a model conditioned on scene descriptions, \eg, text prompts, box maps, etc., to generate scene videos for $M$ modalities, \eg, RGB videos, depth map videos, semantic map videos, etc.
Each scene video consists of $K$ frames and $V$ camera views.

\vspace{-0.5cm}
\paragraph{Pipeline Overview.}
Fig.~\ref{fig_framework} shows an overview of our approach.
Our approach follows the paradigm of diffusion-based video generation~\cite{gao2025magicdrive-v2,wang2024drivedreamer}.
As shown in the left of Fig.~\ref{fig_framework}, we construct scene descriptions with text conditions, context reference conditions, and layout conditions, and employ the corresponding encoders to extract the embeddings of these conditions.
The embeddings of layout conditions and context reference conditions are concatenated with noisy latents $x$ as inputs $z$ to a diffusion model, while the embeddings of text conditions are injected into the diffusion model through cross-attention layers.
Here, noisy latents $x$ are obtained by sampling random noises from the Gaussian distribution during inference and by adding scheduled noises to the groundtruth frame latents extracted by a VAE encoder during training.
Next, as shown in the middle of Fig.~\ref{fig_framework}, we construct a multi-modal multi-view diffusion transformer model with modal-shared layers, composed of temporal layers and multi-view spatiotemporal blocks, and modal-specific layers, composed of cross-modal interaction layers and projection layers.
We then train the diffusion transformer model to estimate the added noises $\epsilon{\sim}\mathcal{N}(0,I)$ from the conditioned inputs $z$, so as to generate clean latents $x'$ for each modality.
Finally, as shown in the right of Fig.~\ref{fig_framework}, the clean latents $x'$ are used as inputs to a VAE decoder to produce multi-view scene videos for each modality.
In the following sections, we present the details of conditioning inputs encoding, multi-modal multi-view diffusion transformer, and model training and inference.

\subsection{Conditioning Inputs Encoding}

\paragraph{Text Conditions.}
Following~\cite{gao2025magicdrive-v2,wang2024drivedreamer}, text conditions are used to guide holistic scene generation.
In our framework, we employ two types of text conditions, including camera prompts and text prompts, and use two text encoders to generate text condition embeddings $f^{text}$.
For camera prompts $c^{cam}$, we concatenate the camera intrinsic and extrinsic parameters applied with the Fourier embedding~\cite{mildenhall2021nerf} and employ an MLP-based encoder $E^{cam}$ to extract camera embeddings.
For text prompts $c^{text}$, we use video caption descriptions and apply a frozen text encoder $E^{text}$ to extract text embeddings.
We concatenate these embeddings as $f^{text}$.
This process can be formulated as:
\begin{equation}
    \label{eq:text_condition}
    f^{text} = E^{cam}(c^{cam}) \otimes E^{text}(c^{text}),
\end{equation}
where $\otimes$ denotes concatenation.

\vspace{-0.5cm}
\paragraph{Layout Conditions.}
For fine-grained scene structure and content control, we employ three types of layout conditions, including box maps, road maps and occupancy-based layout maps.
These layout conditions can be obtained from human annotations or pre-trained models and are useful to control fine-grained details in complex multi-view driving scene generation.
For box maps, different from~\cite{gao2024magicdrive,gao2025magicdrive-v2} which use box coordinates to extract box embeddings, we directly generate 2D box maps $c^{b}$ by projecting target 3D box locations onto image planes and assign different colors to different categories.
For road maps $c^{r}$, we project road structure cues (lane dividers, pedestrian crossings, and lane boundaries) onto image planes and assign different colors to different categories.
In addition, following~\cite{li2024syntheocc}, we use occupancy-based layout maps $c^{o}$ to further enhance scene fine-grained control.
This is accomplished by projecting sparse 3D occupancy onto an image plane to generate sparse semantic maps.
Note that these semantic maps are coarse and sparse, which are different from the dense semantic modality used for urban scene understanding in autonomous driving.
3D occupancy has been well explored in autonomous driving, and in practice, 3D occupancy can be obtained from an off-the-shelf model~\cite{li2025uniscene} or a simulator~\cite{dosovitskiy2017carla}.

To encode these layout conditions, unlike prior work~\cite{wang2024drivedreamer,zhao2025drivedreamer,gao2024magicdrive,gao2025magicdrive-v2} which employs multiple encoders, we propose to use a unified layout encoder to fuse these conditions before feeding them to the diffusion model.
Specifically, we employ causal convolutional layers to construct causal resnet blocks inspired by~\cite{yang2025cogvideox}.
We then construct the unified layout encoder with separate causal blocks for each condition and a shared causal resnet block for all conditions.
In this way, we fuse layout conditions to generate layout condition embeddings $f^{layout}$ for controllable scene generation.
This can be formulated as:
\begin{equation}
    \label{eq:layout_condition}
    f^{layout} = E^{l}_s(E^{l}_b(c^{b}) \otimes E^{l}_r(c^{r}) \otimes E^{l}_o(c^{o})),
\end{equation}
where $E^{l}_b$, $E^{l}_r$ and $E^{l}_o$ are causal resnet blocks for each condition, while $E^{l}_s$ is the shared causal resnet block.

\vspace{-0.5cm}
\paragraph{Context Reference Conditions.}
In urban scene generation, context reference conditions $c^{ref}$ refer to the initial frame conditions which are optional and only used for future scene prediction similar to world models~\cite{wang2024drivedreamer,ni2025maskgwm}.
We employ a 3D VAE encoder $E^{vae}$~\cite{yang2025cogvideox} to encode these context conditions but set the temporal dimension to one since each modality usually has only one initial frame for each camera view.
In our approach, we employ a shared pre-trained 3D VAE from CogVideoX~\cite{yang2025cogvideox} for all modalities, rather than using different VAEs for different modalities.
Since the 3D VAE in CogVideoX~\cite{yang2025cogvideox} is trained with a massive corpus of real-world videos, it performs well for encoding and decoding videos of different modalities.
During model training, we use the shared pre-trained 3D VAE encoder to extract groundtruth frames latents to generate noisy latents.
Therefore, the conditioned inputs $z$ is obtained by:
\begin{equation}
    \label{eq:reference}
    z = E^c(f^{layout} \otimes E^{vae}(c^{ref})) \otimes x,
\end{equation}
where $E^c$ is a convolutional layer to modulate the embedding dimension to be the same as $x$.

\begin{figure}[t]
\centering
\includegraphics[width=0.68\columnwidth]{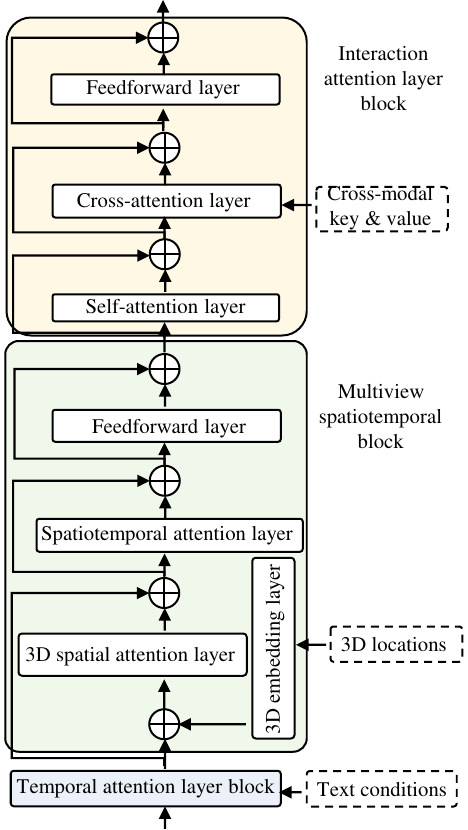}
\vspace{-0.3cm}
\caption{An illustration of our diffusion transformer block.
}
\label{fig_architecture}
\end{figure}

\subsection{Multi-modal Multi-view Diffusion Transformer}
\label{sec:transformer}
Contemporary video diffusion models cannot be directly used for multi-view controllable urban scene generation.
On the other hand, existing driving video diffusion models lack the ability to generate multi-modal data for holistic scene understanding.
To resolve this problem, we devise a new multi-modal multi-view diffusion transformer model.
Specifically, since a shared pre-trained 3D VAE $E^{vae}$ can be used to encode and decode videos of different modalities, we conjecture that \emph{different modalities share a common latent space} and only require certain components to capture modal-specific content in order to distinguish them.
In light of this, we decompose the model learning process into modal-shared learning and modal-specific learning and construct a unified diffusion transformer model with modal-shared components and modal-specific components.

\vspace{-0.5cm}
\paragraph{Modal-Shared Components.}
We use temporal attention layers $D^{tem}$ with 3D full attention from CogVideoX~\cite{yang2025cogvideox} to learn temporal consistency across video frames and inject text conditions $f^{text}$ through cross-attention in the temporal attention layers.
Here, the dimension of conditioned inputs $z$ of each modality is transformed from $\mathcal{R}^{V{\times}K{\times}H{\times}W{\times}C}$ to $\mathcal{R}^{V{\times}(NKW){\times}C}$ for the temporal attention layers.
However, these temporal layers cannot guarantee spatiotemporal consistency for multi-view video generation.
Hence, we append multi-view spatiotemporal blocks $D^{st}$ after temporal layers blocks to learn scene structure and capture spatiotemporal consistency.
As shown in Fig.~\ref{fig_architecture}, each multi-view spatiotemporal block consists of a 3D spatial attention layer, a 3D spatial embedding layer, a spatiotemporal attention layer and a feedforward layer.
The 3D spatial embedding layer is a multi-resolutional Hash grid~\cite{muller2022instant} that encodes 3D occupancy locations $c^{occ}$ into 3D spatial embeddings, which are added to latents from the temporal layers block to enhance spatial consistency.
The 3D spatial attention layer is a self-attention layer that transforms the latent dimension to $\mathcal{R}^{K{\times}(VHW){\times}C}$ to learn 3D spatial structure information of all surrounding camera views.
The spatiotemporal attention layer is a 3D full attention layer that transforms the latent dimension to $\mathcal{R}^{(VKHW){\times}C}$ to capture full spatiotemporal information for multi-view driving scene generation.
In practice, we append the multi-view spatiotemporal block after every $\alpha_1$ temporal layers block rather than every block.
The feedforward layer is a fully-connected layer used to further transform latents.
Following~\cite{yang2025cogvideox}, we also employ adaptive normalization between each layer and adaptive scaling and shifting to modulate latents.
We define this process as:
\begin{equation}
    \label{eq:modal_shared}
    h = D^{st}(D^{tem}(z', f^{text}, t), c^{occ}, t),
\end{equation}
where $t$ is the diffusion timestep, $z'$ is the latents output from the previous layers, and $h$ is the latents output from the modal-shared layers.

\vspace{-0.5cm}
\paragraph{Modal-Specific Components.}
We construct each cross-modal interaction layer block $D^{cm}_m$ with a self-attention layer, a cross-attention layer, and a feedforward layer, where adaptive normalization between each layer and scaling and shifting are also used.
Here, the query of the cross-attention layer is the latents of each modality, while the key and the value are from the concatenated latents of the other modalities.
This cross-modal interaction layer block is used to learn modal-specific content while maintaining cross-modal alignment.
In practice, the cross-modal interaction layer block is appended after every $\alpha_2$ modal-shared layers.
This process for each modality is defined as:
\begin{equation}
    \label{eq:modal_specific}
    h'_m = D^{cm}_m(h, h^{modal}_m, t),
\end{equation}
where $h'_m$ is the modal-specific latents output from the cross-modal interaction layer, $h^{modal}_m$ denotes the key and the value from other modalities for the cross-attention layer.
After $N$ modal-shared layers and cross-modal interaction layers, we employ the modal-specific projection head (a linear layer with adaptive normalization) to estimate the added noises $\epsilon$ and generate the clean latents $x'$ for each modality.
The clean latents $x'$ for each modality is used as the input to a VAE decoder~\cite{yang2025cogvideox} to generate modal-specific multi-view scene videos.
In this way, our diffusion transformer model can generate multi-modal multi-view videos with high controllability and spatiotemporal consistency in a unified framework. 
Our model can be used for both single-modal and multi-modal generation, and modalities are not limited to RGB videos, depth maps, and semantic maps.

\subsection{Model Training and Inference}
For model training, we employ the DDPM~\cite{ho2020denoising} noise scheduler and define the training objective $\mathcal{L}$ as:
\begin{equation}
    \label{eq:loss}
    \mathcal{L} = \sum_{m}^{M}{\lambda_m}{\mathbb{E}_{x_{0,m}, t_m, \epsilon_m, C}{\parallel}\epsilon_m{-}\epsilon_{\theta,m}(x_{t,m}, t_m, C){\parallel}^2},
\end{equation}
where for the $m$th modality, $\lambda_m$ is the weight for the training loss, $x_{0,m}$ is the groundtruth clean latent, $x_{t,m}$ is the noisy latent, $C$ is the condition, $\epsilon_{\theta,m}$ is the diffusion model, $t_m$ is the timestep.
Besides, conditioning dropout is used as an augmentation strategy to improve model generalization and output diversity.
During inference, random noises are sampled from a Gaussian distribution as the inputs.
We employ the DDIM~\cite{song2021denoising} sampler to improve the efficiency of the reverse diffusion process and classifier-free guidance~\cite{ho2022classifier} to balance output diversity and condition consistency.

\begin{table*}[t!]
    \small
    \centering
    \caption{Quantitative comparison with the state-of-the-arts on nuScenes.
    Some results are cited from the original papers.
    $^{\star}$The ``diverse conditions'' adopted may vary across methods.
    $^{\ddagger}$Implemented with cross-view and cross-frame attention.
    }
    \begin{tabular}{l|l|c|c|c|cc|c|c}
        \hline
        \multirow{2}{*}{Method} & \multirow{2}{*}{Venue}
        & \multicolumn{1}{c|}{Diverse$^{\star}$} & \multirow{2}{*}{Multi-modal?} 
         & \multicolumn{1}{c|}{Fidelity} & \multicolumn{2}{c|}{Controllability}
         & \multicolumn{1}{c|}{Depth quality} & \multicolumn{1}{c}{Semantic quality} \\
         & & conditions? &  & FVD$\downarrow$ & mAP$\uparrow$ & mIoU$\uparrow$ & AbsRel$\downarrow$ & mIoU$\uparrow$ \\
        \hline
        \hline                
        DriveDreamer~\cite{wang2024drivedreamer} & ECCV24   & \cmark & \xmark  &  340.8    & -         & -     & -         & -     \\
        DrivingDiffusion\cite{li2024drivingdiffusion}& ECCV24& \cmark & \xmark  & 332.0    & -         & -     & -         & -     \\
        Drive-WM~\cite{wang2024driving}  &CVPR24      & \cmark & \xmark  &  122.7    & -         & -     & -         & -     \\
        Panacea~\cite{wen2024panacea} &CVPR24        & \cmark & \xmark  &  139.0    & -         & -     & -         & -     \\
        MagicDrive~\cite{gao2024magicdrive} & ICLR24     & \cmark & \xmark  &  236.2    & 9.7      & 15.6  & 0.255 & 23.5  \\
        DriveDreamer-2~\cite{zhao2025drivedreamer} & AAAI25 & \cmark & \xmark  &  {55.7}     & -         & -     & -         & -     \\                       
        UniScene~\cite{li2025uniscene}  & CVPR25      & \cmark & \xmark  &  70.5     & -         & -     & -         & -     \\
        MaskGWM~\cite{ni2025maskgwm}  & CVPR25       & \cmark & \xmark  &  65.4     & -         & -     & -         & -     \\
        MagicDrive-V2~\cite{gao2025magicdrive-v2} & ICCV25  & \cmark & \xmark   &  112.7     & 11.5      & 17.4  & 0.280  & 22.4  \\
        \hline
        CogVideoX~\cite{yang2025cogvideox} + & ICLR25+    & \multirow{2}{*}{\cmark} & \multirow{2}{*}{\xmark}  & \multirow{2}{*}{60.4} &  \multirow{2}{*}{15.9}  & \multirow{2}{*}{28.2} & \multirow{2}{*}{0.124} & \multirow{2}{*}{32.4} \\
        SyntheOcc$^{\ddagger}$~\cite{li2024syntheocc}&arXiv24&&&&&&\\
        \hline       
        MoVieDrive (Ours)  & -         & \cmark & \cmark   &  \textbf{46.8}     & \textbf{22.7}      & \textbf{35.8}    & \textbf{0.110}   & \textbf{37.5}   \\ 
        \hline
    \end{tabular}    
    \label{tab:exp-sota}
\end{table*}

\section{Experiments}
\label{sec:experiments}

\subsection{Dataset and Experimental Setup}
\label{sec:dataset}
\paragraph{Dataset.}
Following the state-of-the-arts~\cite{gao2024magicdrive,gao2025magicdrive-v2,wang2024drivedreamer,zhao2025drivedreamer}, we mainly conduct experiments on the nuScenes~\cite{caesar2020nuscenes} dataset, a real-world autonomous driving benchmark.
On nuScenes, we use the official training and validation splits.
For each training scene, we randomly select a sequence at each iteration.
For the layout condition, we extract road maps and box maps following~\cite{wang2024drivedreamer,gao2024magicdrive} and prepare 3D occupancies following~\cite{li2024syntheocc}.
For context reference conditions, we employ the first frames of all camera views following the common practice~\cite{zhao2025drivedreamer}.
For the text conditions, we utilize the camera intrinsic and extrinsic parameters as the camera prompt and prepare text prompts using CogVLM2-Caption~\cite{yang2025cogvideox}.
For multi-modal data preparation, since no groundtruth multi-modal data are provided, following the common practice, we generate depth maps with Depth-Anything-V2~\cite{yang2024depth} and semantic maps with Mask2Former~\cite{cheng2022masked} for multi-modal training.
In practice, when groundtruth multi-modal data are provided, our approach can be used to train a unified model for multi-modal multi-view scene generation.
Here, we use these models because they are widely used for depth and semantic prediction in autonomous driving scene understanding, but other off-the-shelf models can also be used to prepare multi-modal data, and we leave the integration of more state-of-the-art models for future work.
Additionally, we evaluate our approach on the Waymo dataset~\cite{sun2020scalability}.
On Waymo, we use the official training and validation splits with the three front views for video generation.

\vspace{-0.5cm}
\paragraph{Implementation Details.}
We build our model on CogVideoX(v1.1-2B)~\cite{yang2025cogvideox} and SyntheOcc~\cite{li2024syntheocc}.
The 3D VAE encoder and decoder are from~\cite{yang2025cogvideox} and initialized with the pre-trained weights.
For text condition encoding, we employ the pre-trained T-5~\cite{raffel2020exploring} as the frozen $E^{text}$
and construct $E^{cam}$ with an MLP-based encoder consists of two linear layers, where LayerNorm and SiLU are used between layers.
For layout condition encoding, we use causal convolutional layers to construct causal resnet blocks following~\cite{yang2025cogvideox} and then construct the unified layout encoder with separate causal blocks for each condition and a shared causal resnet block for all conditions.
For context reference condition encoding, we use the frozen 3D VAE as $E^{vae}$ but set the temporal dimension to one.
We construct the multi-modal multi-view diffusion transformer with the architecture presented in Section~\ref{sec:transformer}, where the temporal layers and projection heads are initialized with the pre-trained weights from~\cite{yang2025cogvideox} and the other layers are randomly initialized.
During training, the 3D VAE and the T5 text encoder are frozen, while the other model components are jointly trained.
We train our model with the AdamW optimizer using a learning rate of $2e^{-4}$.
By default, we use $6$ cameras and $49$ video frames, each with a resolution of $512{\times}256$.

\subsection{Driving Scene Video Generation}
\paragraph{Evaluation Metrics.}
For video fidelity evaluation, we use the commonly used Fr\'{e}chet Video Distance (FVD)~\cite{unterthiner2018towards} as the metric.
To evaluate the controllability, following~\cite{gao2025magicdrive-v2}, we use the pre-trained BEVFormer~\cite{li2022bevformer} to evaluate video-based 3D object detection in terms of mAP and video-based BEV segmentation in terms of mIoU.

\vspace{-0.5cm}
\paragraph{Compared Methods.}
We compare our approach with state-of-the-art urban scene video generation methods, including MagicDrive~\cite{gao2024magicdrive}, MagicDrive-V2~\cite{gao2025magicdrive-v2}, DriveDreamer~\cite{wang2024drivedreamer}, DriveDreamer-2~\cite{zhao2025drivedreamer}, Drive-WM~\cite{wang2024driving}, Panacea~\cite{wen2024panacea}, UniScene~\cite{li2025uniscene}, MaskGWM~\cite{ni2025maskgwm}, and DrivingDiffusion~\cite{li2024drivingdiffusion}.
Here, different methods propose to use different diverse conditions, so we mainly compare with the original methods instead of modifying them.
Besides, we report results of CogVideoX~\cite{yang2025cogvideox} + SyntheOcc~\cite{li2024syntheocc}(with cross-view attention layers) for multi-view scene video generation and consider this model as our direct competitor.
Note that single-view methods~\cite{gao2024vista} are not our direct competitors since they cannot be directly used for multi-view video generation.

\begin{figure}[t!]
\centering
\includegraphics[width=0.99\columnwidth]{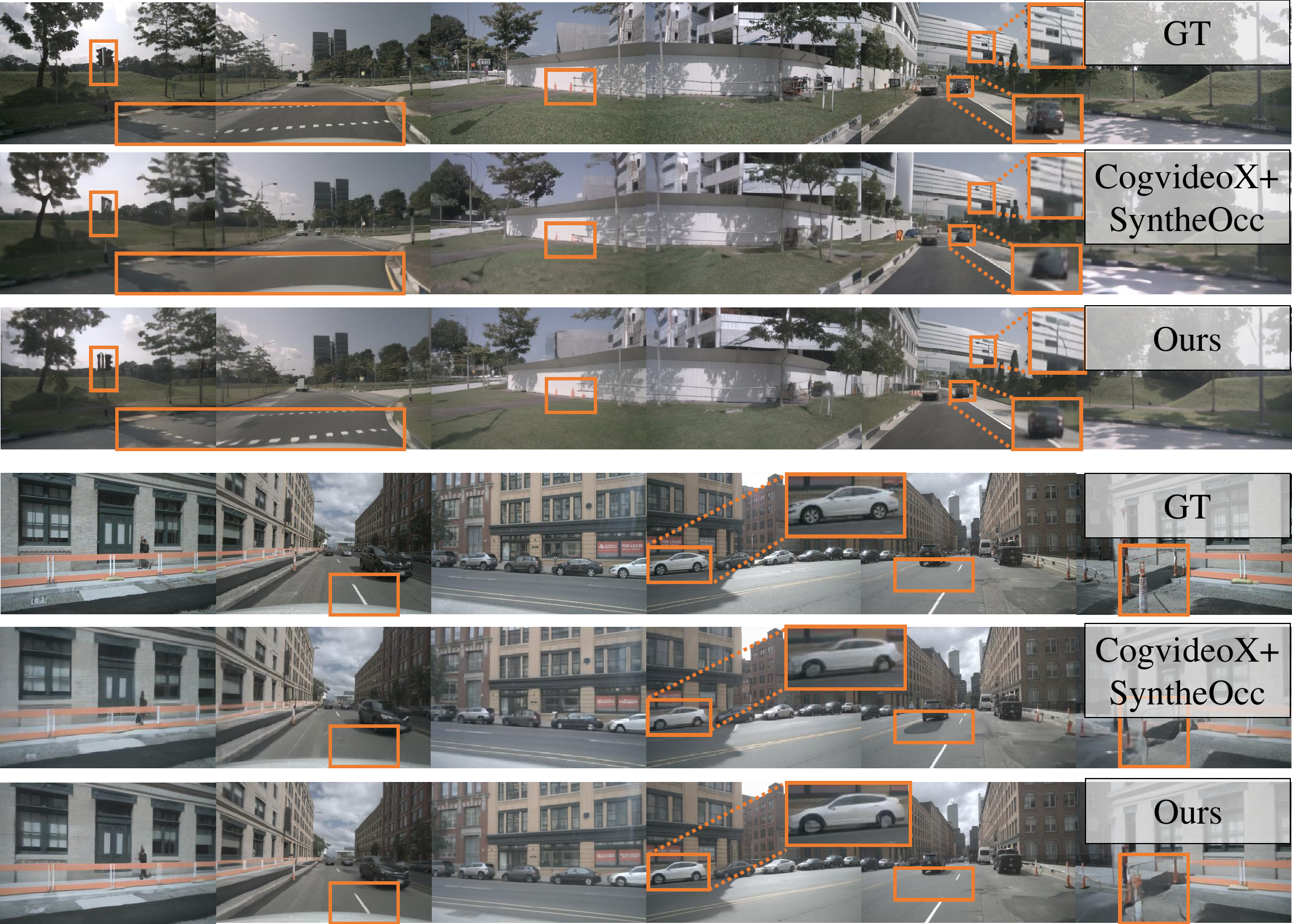}
\vspace{-0.3cm}
\caption{Quantitative comparison on nuScenes.
From left to right, we show results of back-right, back, back-left, front-left, front and front-right cameras and highlight some noticeable details.
Please check the supplementary material for an enlarged version.
}
\label{fig_sota}
\end{figure}

\vspace{-0.5cm}
\paragraph{Result Analysis of Video Fidelity.}
Video fidelity measures the realism and temporal coherence of generated urban scene videos.
We report FVD results in Tab.~\ref{tab:exp-sota}.
We can see that our approach achieves better results compared with state-of-the-art methods.
Specifically, our approach achieves FVD of 46.8, which is better than DriveDreamer, DriveDreamer-2, UniScene, MaskGWM, MagicDrive, etc.
Besides, compared with CogVideoX + SyntheOcc, our approach improves performance by around 22\%.
Fig.~\ref{fig_sota} shows some qualitative comparison with state-of-the-art methods.
From Fig.~\ref{fig_sota}, we can see that our approach can produce urban scene videos with high-fidelity details, \eg, vehicles and road structures.

\vspace{-0.5cm}
\paragraph{Result Analysis of Scene Controllability.}
Scene controllability refers to the consistency between the conditioning inputs and the generated videos.
We report mAP for 3D object detection and mIoU for BEV segmentation in Tab.~\ref{tab:exp-sota}.
We can see that our approach achieves the best mAP of 22.7 for 3D object detection and the best mIoU of 35.8 for BEV segmentation, outperforming state-of-the-art methods.
This indicates that our approach can generate urban scene videos that are more consistent with the control conditions.

\begin{figure}[t!]
\centering
\includegraphics[width=0.99\columnwidth]{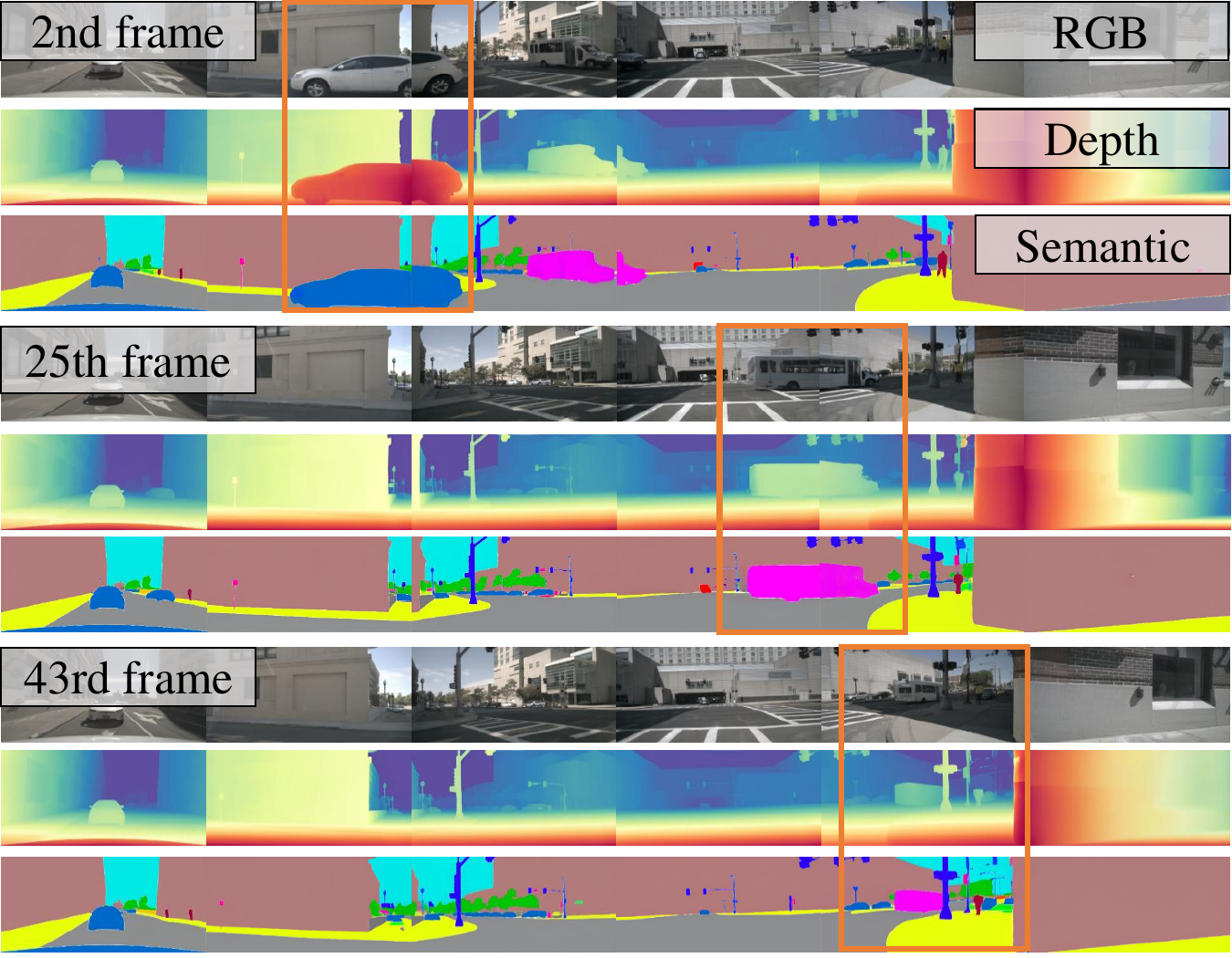}
\vspace{-0.3cm}
\caption{Visualization of cross-modal consistency on nuScenes.}
\label{fig_exp_modal}
\end{figure}

\subsection{Multi-modal Multi-view Video Generation}
The purpose of this experiment is to verify the superiority of using a unified model for multi-modal multi-view video generation compared to the common pipeline of using multiple models to generate different modalities.
Our approach can simultaneously generate RGB videos, depth maps, and semantic maps using the proposed unified framework.
Since existing driving scene video generation methods lack the multi-modal generation capability, we generate the RGB videos first and then use off-the-shelf models to estimate the depth maps and semantic maps.

\vspace{-0.5cm}
\paragraph{Evaluation Metrics.}
To evaluate the quality of the generated depth maps, we use the Absolute Relative Error (AbsRel)~\cite{yang2024depth}.
And to evaluate the quality of the generated semantic maps, we report the mean Intersection over Union (mIoU) for traffic-related classes, buildings, trees, and sky.
The smaller AbsRel and the larger mIoU indicate the better quality of the generated depth and semantic maps.

\vspace{-0.5cm}
\paragraph{Evaluation on Depth Quality and Semantic Quality.}
In Tab.~\ref{tab:exp-sota}, we report the depth quality in terms of AbsRel and semantic quality in terms of mIoU.
We can see that compared with the depth maps generated using a multi-stage pipeline, our approach achieves the best AbsRel of 0.110 and mIoU of 37.5.
This shows that the multi-modal generation in our unified framework yields high-quality depth maps and semantic maps, even without using additional models for multi-modal generation.
In addition, we visualize some multi-modal generation results in Fig.~\ref{fig_exp_modal}.
These results further show the cross-modal consistency of the multi-modal data generated from our approach.

\begin{table}[t!]
    \small
    \centering
    \caption{Ablation study of multi-modal generation on nuScenes.
    `\cmark': using our approach. `\xmark': using the off-the-shelf models.
    }
    \vspace{-0.3cm}
    \begin{tabular}{ccc|c|c|c}
        \hline
        \multicolumn{3}{c|}{Methods} & {RGB} & {Depth} & {Semantic} \\
        RGB & Depth & Semantic &  FVD$\downarrow$ & AbsRel$\downarrow$& mIoU$\uparrow$\\
        \hline
        \hline
        \cmark & \xmark & \xmark  & \textbf{42.0} & 0.121 & 36.4   \\
        \cmark & \cmark & \xmark  & 43.4 & 0.111 & 36.0    \\ 
        \cmark & \cmark & \cmark  & 46.8 & \textbf{0.110} & \textbf{37.5}     \\
        \hline
    \end{tabular}
    \label{tab:exp-ablation-multimodal}
\end{table}

\begin{table}[t!]
    \small
    \centering
    \caption{Effectiveness analysis of the diffusion transformer components on nuScenes.
    `L1': temporal layers, `L2': multi-view spatiotemporal blocks, `L3': modal-specific layers.
    }
    \vspace{-0.3cm}
    \begin{tabular}{ccc|c}
        \hline
        \multicolumn{3}{c|}{Methods} & {RGB} \\
        L1 blocks & L2 blocks & L3 blocks &  FVD$\downarrow$ \\
        \hline
        \hline
        \cmark & \xmark & \xmark  & 153.7 \\
        \cmark & \xmark & \cmark  & 78.8 \\ 
        \cmark & \cmark & \cmark  & \textbf{46.8} \\
        \hline
    \end{tabular}
    \label{tab:exp-ablation-diffusion}
\end{table}

\begin{figure}[t!]
\centering
\includegraphics[width=0.99\columnwidth]{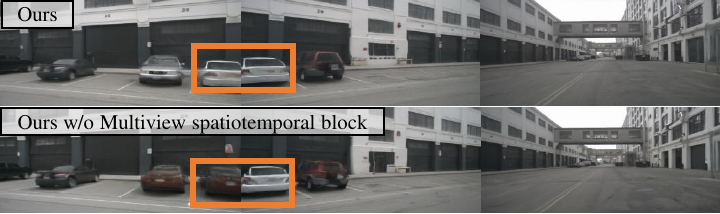}
\vspace{-0.3cm}
\caption{Visualization of cross-view consistency on nuScenes.
}
\label{fig_ablation}
\end{figure}

\subsection{Ablation Study}
\label{sec:ablation}
\paragraph{Effectiveness of Multi-Modal Generation.}
In Tab.~\ref{tab:exp-ablation-multimodal}, we analyze the effectiveness of our multi-modal components.
We compare the performance of ours(RGB) + Depth-Anything-V2 + Mask2Former, ours(RGB + depth) + Mask2Former, and ours(RGB + depth + semantic).
Tab.~\ref{tab:exp-ablation-multimodal} shows that our unified approach (RGB + depth + semantic) achieves the best depth AbsRel and semantic mIoU and comparable RGB FVD.
This verifies that generating multiple modalities in a unified framework can reduce the number of models and achieve better overall performance.

\vspace{-0.5cm}
\paragraph{Effectiveness of Diffusion Transformer Components.}
In Tab.~\ref{tab:exp-ablation-diffusion}, we study the effectiveness of the proposed diffusion transformer model.
We can see that using only temporal layer blocks (`L1') results in poor performance, while using temporal layer blocks plus modal-specific layer blocks (`L1+L3') brings better performance.
With all components (`L1+L2+L3'), our model yields the best performance.
Besides, Fig.~\ref{fig_ablation} shows that our approach maintains good cross-view consistency, while removing the multi-view spatiotemporal blocks results in poor consistency.

\begin{figure}[t!]
\centering
\includegraphics[width=0.9\columnwidth]{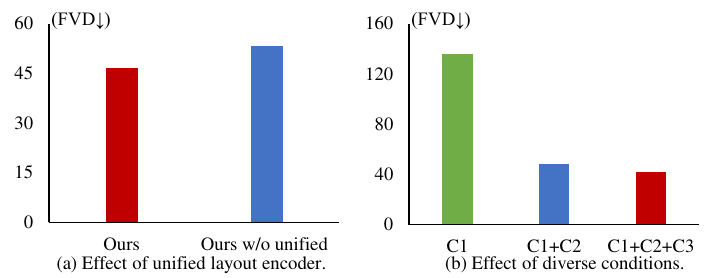}
\vspace{-0.3cm}
\caption{Effectiveness of (a) unified layout condition encoder and (b) diverse conditioning inputs.
`C1': text conditions, `C2': box and road map conditions, `C3': occupancy conditions.
}
\label{fig_exp_encoder_cond}
\end{figure}

\begin{figure}[t]
\centering
\includegraphics[width=0.85\columnwidth]{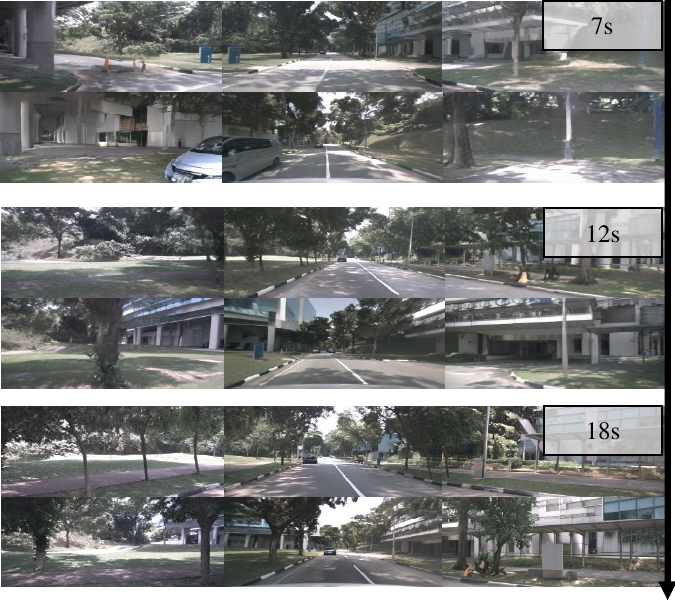}
\vspace{-0.3cm}
\caption{Long video generation without reference frames.
}
\label{fig_exp_long}
\end{figure}

\subsection{Further Analysis and Discussion}
\label{sec:further_analysis}

\paragraph{Effectiveness of Unified Layout Condition Encoder.}
In Fig.~\ref{fig_exp_encoder_cond}(a), we analyze the effectiveness of our unified layout condition encoder.
Here, ours without the unified encoder uses a pre-trained 3D VAE encoder for encoding.
We can see that using the proposed unified layout condition encoder brings better performance compared with using pre-trained 3D VAE encoders.
This can be attributed to the implicit condition embedding space alignment using the proposed module, which effectively fuses the conditions.

\vspace{-0.5cm}
\paragraph{Effectiveness of Diverse Conditioning Inputs.}
In Fig.~\ref{fig_exp_encoder_cond}(b), we examine the effectiveness of the diverse conditions.
We can see that only using text prompt results in poor performance, while adding box and road map conditions brings significant performance improvement, which is aligned with existing work.
Besides, the use of occupancy conditions also improves performance.

\vspace{-0.5cm}
\paragraph{Long Video Generation without Reference Frames.}
Although our approach is not specifically designed for long video generation, our model can be extended for long video generation, even without reference frames.
This is benefited from the temporal compression mechanism in 3D VAE~\cite{yang2025cogvideox}, the proposed diffusion transformer model and the diverse conditions.
We visualize some results in Figs.~\ref{fig_illustration}(c) and~\ref{fig_exp_long}.
We can see that our approach can generate long videos with consistent scene layout and content.
Nevertheless, we also observe the presence of noisy regions in far-distance areas.

\begin{figure}[t!]
\centering
\includegraphics[width=0.9\columnwidth]{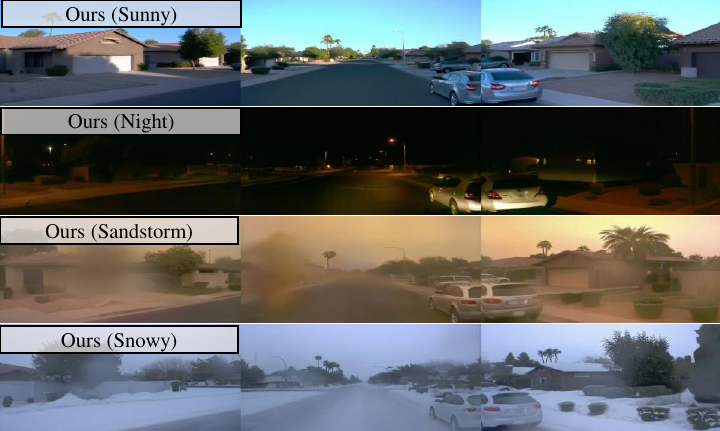}
\vspace{-0.3cm}
\caption{Visualization of experimental results on Waymo.
}
\label{fig_exp_waymo}
\end{figure}

\vspace{-0.5cm}
\paragraph{Evaluation on the Waymo dataset.}
Furthermore, we evaluate on the Waymo dataset.
Tab.~\ref{tab:exp-sota-waymo} shows that on Waymo, our approach achieves FVD of 61.6, outperforming the baseline (CogVideoX + Cross-view attention + Our conditions) and the state-of-the-art CogVideoX+SyntheOcc.
As shown in Fig.~\ref{fig_exp_waymo}, our approach can generate diverse driving scenes with high-fidelity on Waymo.

\begin{table}[t!]
    \small
    \centering
    \caption{Evaluation on Waymo.
    }
    \vspace{-0.3cm}
    \begin{tabular}{l|c}
        \hline
        Methods             & {FVD$\downarrow$ } \\        
        \hline
        \hline
        Ours              & 61.6 \\
        \hline
        CogVideoX + Cross-view attention + Our conditions & 120.6 \\
        \hline
        CogVideoX~\cite{yang2025cogvideox} + SyntheOcc~\cite{li2024syntheocc} & 82.3 \\
        \hline
    \end{tabular}
    \label{tab:exp-sota-waymo}
\end{table}

\vspace{-0.5cm}
\paragraph{Driving Scene Style Editing with Text Prompts.}
Our approach can also be used to transfer scene style by editing text prompts.
As shown in Figs.~\ref{fig_illustration}(b) and~\ref{fig_exp_waymo}, by editing the scene text prompts, we can generate diverse driving scenes under different time and weather conditions.

\section{Conclusion}
\label{sec:conclution}
This work presents a novel multi-modal multi-view urban scene video generation approach to autonomous driving.
The key idea is to devise a new multi-modal multi-view diffusion transformer model and leverage diverse conditioning inputs to encode controllable scene structure and content cues in a unified framework.
Extensive experiments on two autonomous driving datasets demonstrate the superiority of the proposed approach over state-of-the-art methods.

\clearpage
{
    \small
    \bibliographystyle{ieeenat_fullname}
    \bibliography{main}
}


\clearpage
\setcounter{page}{1}
\maketitlesupplementary

In this supplementary material, we discuss the limitation of this work in Section~\ref{sec:limitation},
present more discussions in Section~\ref{sec:suppl_more_exp},
and present additional visualizations of multi-view multi-modal generation results in Section~\ref{sec:suppl_more_visual}.

\section{Limitation and Future Work}
\label{sec:limitation}
While our approach has achieved superior multi-modal multi-view generation performance, there are still some limitations.
First, how to effectively combine the proposed approach with a closed-loop autonomous driving simulator is worth further study.
This potentially helps to comprehensively assess the safety and reliability of autonomous driving systems.
Second, how to derive LiDAR point clouds from the generated multi-modal data is an interesting research direction to further enhance our approach.
Moreover, as discussed in Section~\ref{sec:further_analysis}, although our approach can be extended for long video generation, there is still room for improvement in terms of long video quality.
Our future work aims to address these problems to facilitate the deployment of our approach for real-world applications.

\section{More Discussions}
\label{sec:suppl_more_exp}

\paragraph{Is this a straightforward modification of CogVideoX with cross-view attention?}
Our approach is not a straightforward modification of CogVideoX~\cite{yang2025cogvideox} for driving scene video generation.
Although we use temporal layers and 3D VAE from CogVideoX, we significantly modify them with a unified multi-modal multi-view diffusion transformer model for autonomous driving scene generation and propose efficient diverse conditioning inputs encoding to improve scene video generation quality and controllability.
A straightforward modification of CogVideoX is adding cross-view attention layers, specific heads, and our conditions to CogVideoX.
However, as shown in Tab.~\ref{tab:suppl_exp_variant}, this yields worse performance compared with our design.
Another simple variant is using the fixed CogVideoX and 3D VAE with learnable modal-specific heads but using neither our multi-view spatiotemporal blocks nor cross-view attention layers.
From Tab.~\ref{tab:suppl_exp_variant}, we can see that this simple variant achieves significantly worse performance.
In addition, combining CogVideoX with SyntheOcc~\cite{li2024syntheocc} still yields worse results compared with our approach.

\begin{table}[t!]
    \small
    \centering
    \caption{Comparison with other model vairants on nuScenes.
    }
    \vspace{-0.3cm}
    \begin{tabular}{l|c}
        \hline
        Methods             & {FVD$\downarrow$ } \\        
        \hline
        \hline
        Ours              & 46.8 \\
        \hline
        CogVideoX + Cross-view attention + Our conditions & 118.4\\
        \hline
        Fixed CogVideoX + Learnable head & 364.1 \\
        \hline
        CogVideoX + SyntheOcc & 60.4 \\
        \hline
    \end{tabular}
    \label{tab:suppl_exp_variant}
\end{table}

\begin{table}[t!]
    \small
    \centering
    \caption{Comparison with a modal-specific variant on nuScenes.
    }
    \vspace{-0.3cm}
    \begin{tabular}{l|c}
        \hline
        Methods                              & {FVD$\downarrow$ } \\        
        \hline
        \hline
        Ours                                 & 46.8 \\
        \hline
        Modality-specific head based variant & 109.4 \\
        \hline
    \end{tabular}
    \label{tab:suppl_exp_variant_head}
\end{table}

\vspace{-0.3cm}
\paragraph{Is this a simple extension of existing multi-modal synthesis methods?}
Our approach is not a simple extension of existing multi-modal synthesis methods.
Compared with existing methods, the innovation of this work lies in that we propose the first work that exploits diverse conditioning inputs and a unified multi-modal multi-view diffusion transformer model for multi-modal multi-view autonomous driving scene generation.
This fills a gap left by existing works, facilitating urban scene understanding in autonomous driving.
As discussed in Section~\ref{sec:related_work}, none of the existing diffusion-based multi-modal synthesis methods are specifically designed for multi-modal multi-view autonomous driving scene generation.
It is not a trivial task to modify these methods for driving scene generation.
For example, HyperHuman~\cite{liu2024hyperhuman} is one of the recent works that explores diffusion-based multi-modal synthesis, but it is specifically devised for human generation so adapting this method to our problem requires modifying conditioning input encoders, diffusion transformer layers, training processes, \etc.
Similar situations applied to other existing methods.
To alleviate this concern, we implement a model variant that employs the modality-specific heads~\cite{xi2025omnivdiff}, multi-view spatiotemporal blocks, CogVideoX, 3D VAE,  and ours conditions.
As shown in Tab.~\ref{tab:suppl_exp_variant_head}, this variant still yields worse performance compared with our approach.
Although more careful modifications of existing multi-modal synthesis methods may improve their performance, from this result we know that adapting existing general methods to multi-modal multi-view driving scene video generation is not a trivial task.
We leave more comprehensive studies for future work.

\begin{figure*}[t]
\centering
\includegraphics[width=0.99\textwidth]{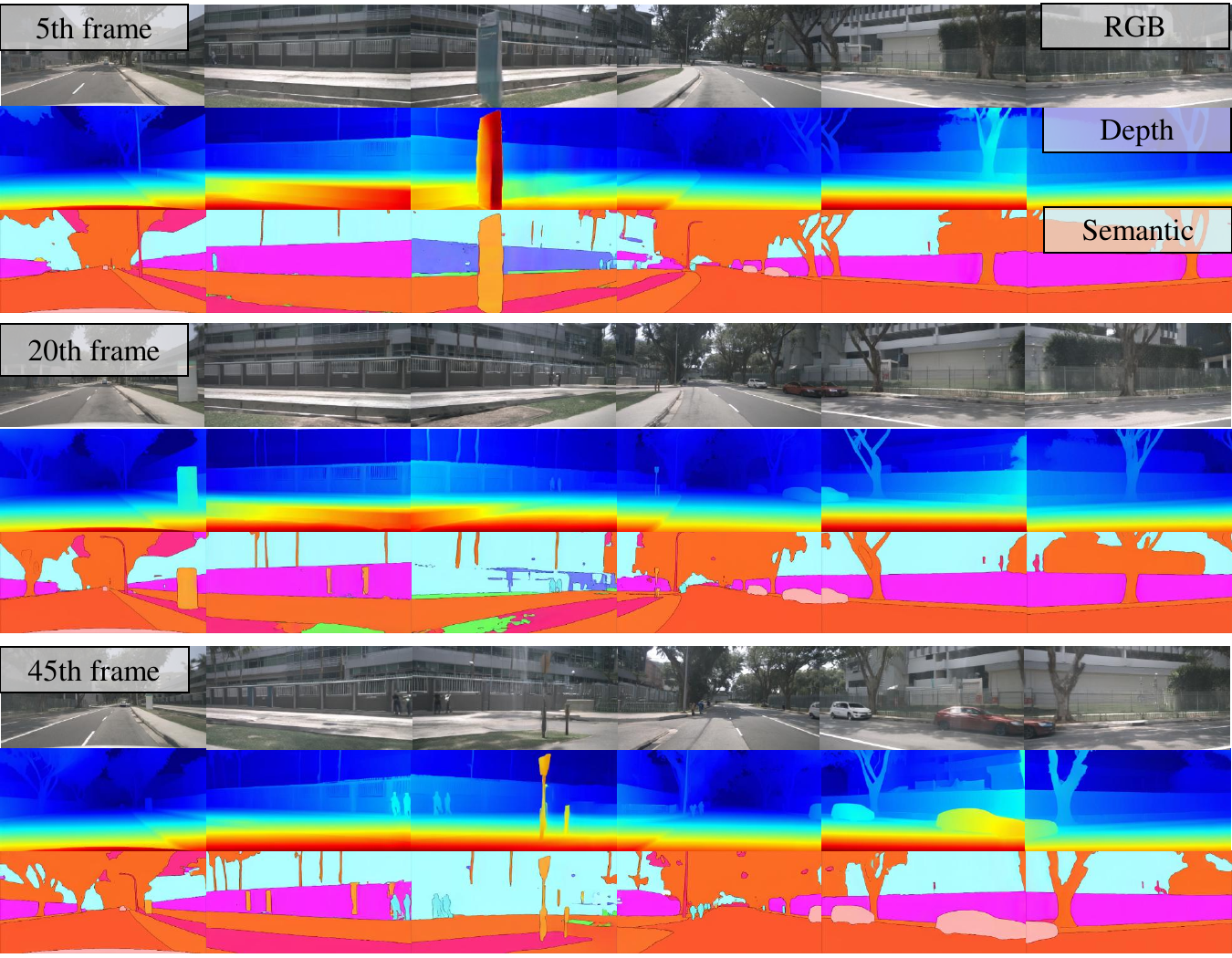}
% \vspace{-0.3cm}
\caption{Multi-modal multi-view generation with DVIS++ and VDA for multi-modal data preparation. 
}
\label{supple_off_the_shelf}
\end{figure*}

\vspace{-0.3cm}
\paragraph{Can this approach be used with other off-the-shelf models for multi-modal data preparation?}
Our approach is orthogonal to the used off-the-shelf models.
In practice, when groundtruth multi-modal data are provided, our approach can be used to train a unified model for multi-modal multi-view scene generation.
Since no groundtruth multi-modal data are provided on the nuScenes and Waymo datasets, we use Depth-Anything-V2 and Mask2Former in this work.
These models are widely used for depth and semantic prediction in autonomous driving scene understanding.
It is worth noting that RGB video generation is key for this task.
If the quality of generated RGB frames is poor, multi-modal prediction can hardly yield good performance.
Thus, the collaboration of multi-modal multi-view generation is a highlight of our approach and Tab.~\ref{tab:exp-ablation-multimodal} verifies that our unified approach can reduce the number of models used and achieve better overall performance for multi-modal multi-view urban scene synthesis.
To alleviate this concern, we further conduct experiments with DVIS++~\cite{zhang2025dvis++} for semantic maps preparation and VDA~\cite{chen2025video} for depth maps preparation.
Experimental results show that our approach with DVIS++ and VDA yields FVD of 47.4, which is similar to the performance of our approach with Depth-Anything-V2 and Mask2Former (FVD of 46.8).
In Fig.~\ref{supple_off_the_shelf}, we show some results of our approach with DVIS++ and VDA.
We can observe that our approach can still generate high-quality multi-modal multi-view urban scene videos with cross-view consistency and cross-modal consistency.
Note that colors of the multi-modal maps are depended on the predefined color palette of the off-the-shelf models and can be mapped to calculate the metrics.

\section{More Visualization Results}
\label{sec:suppl_more_visual}

\begin{figure*}[t]
\centering
\includegraphics[width=0.99\textwidth]{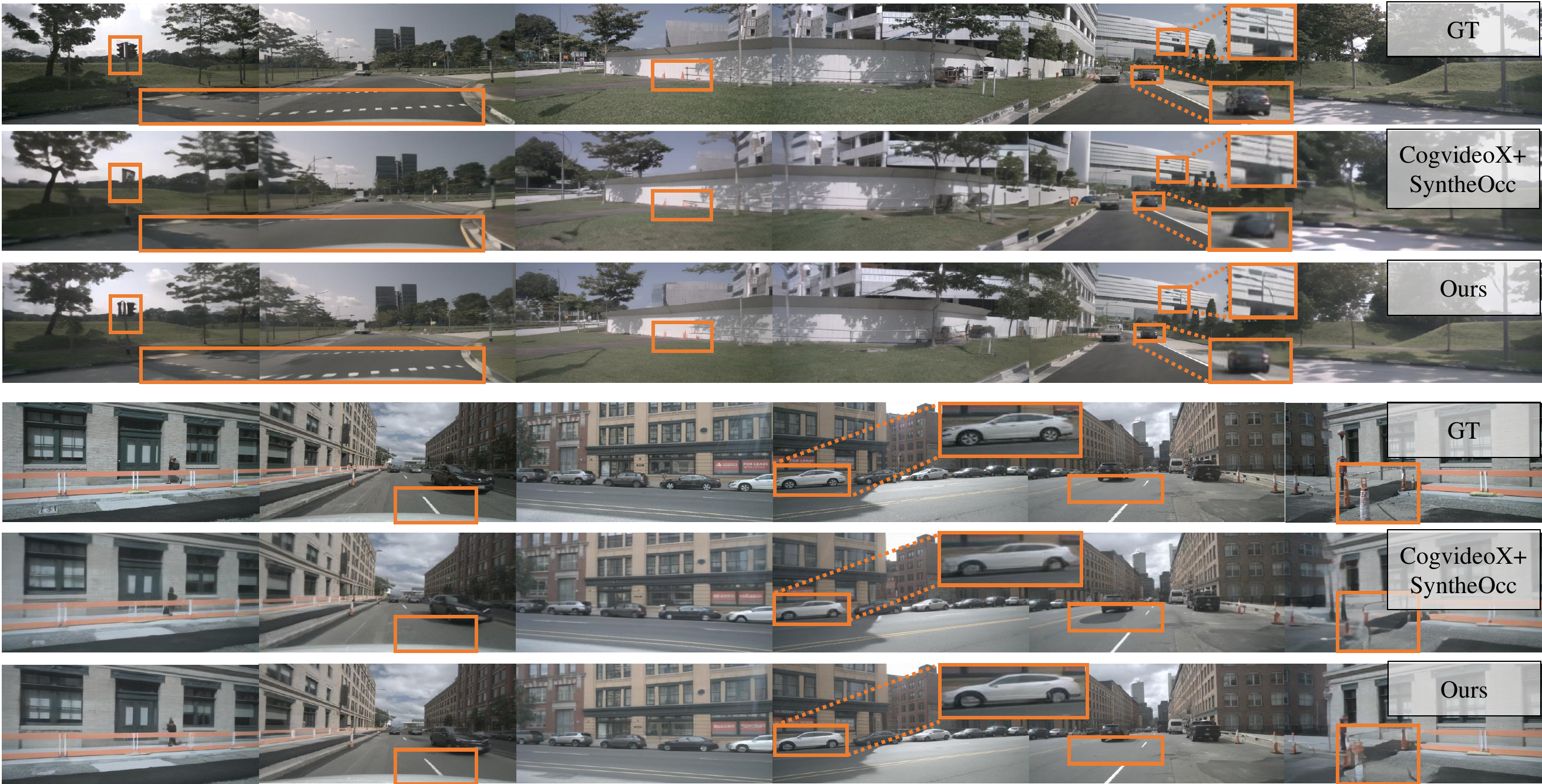}
% \vspace{-0.3cm}
\caption{Quantitative comparison on nuScenes (an enlarged version of Fig.~\ref{fig_sota}).
}
\label{supple_fig_1}
\end{figure*}

\begin{figure*}[t]
\centering
\includegraphics[width=0.99\textwidth]{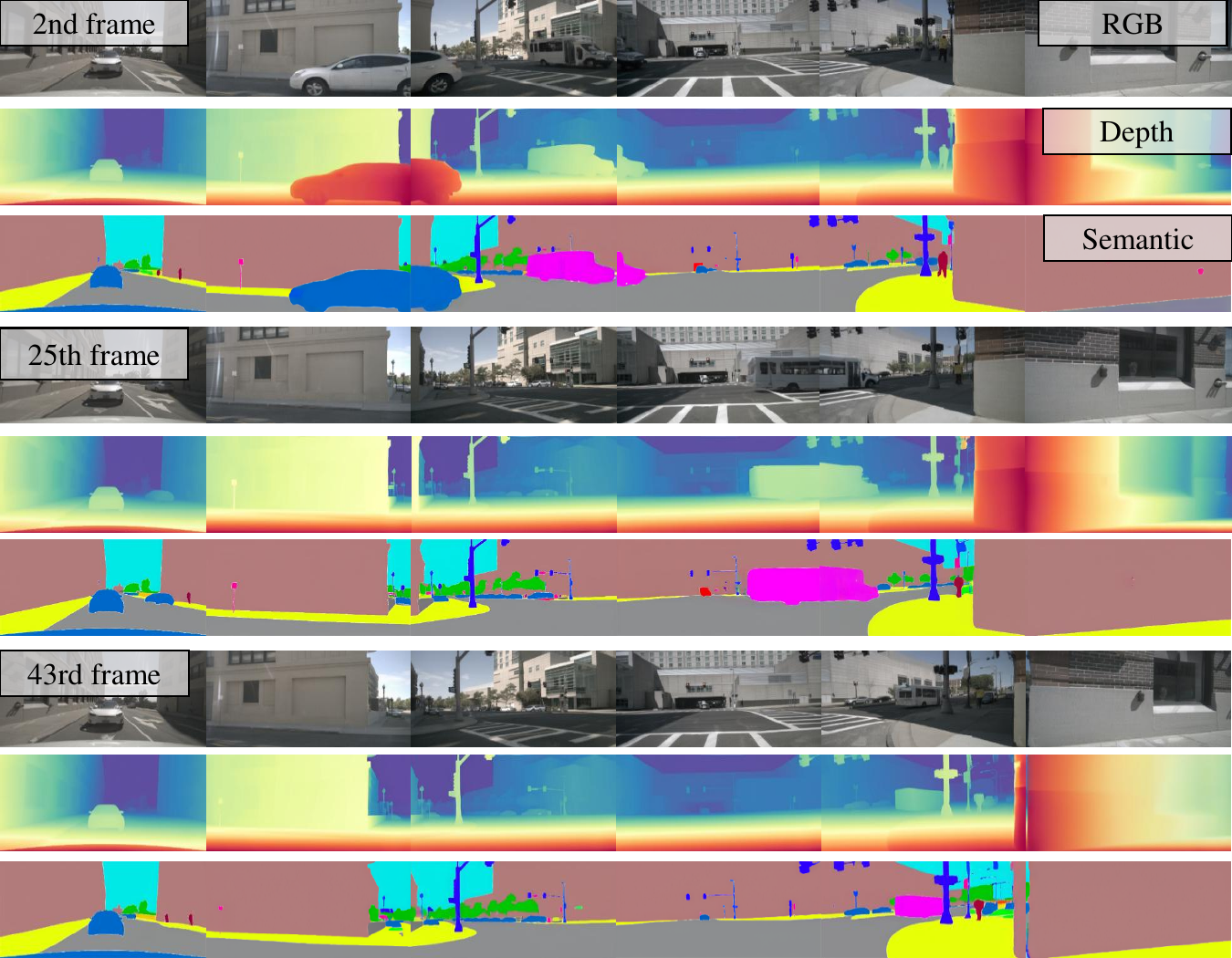}
% \vspace{-0.3cm}
\caption{Additional visualizations of our multi-view multi-modal generation results on nuScenes (arbitrarily selected frames covering different time).
}
\label{supple_fig_2}
\end{figure*}

\begin{figure*}[t]
\centering
\includegraphics[width=0.99\textwidth]{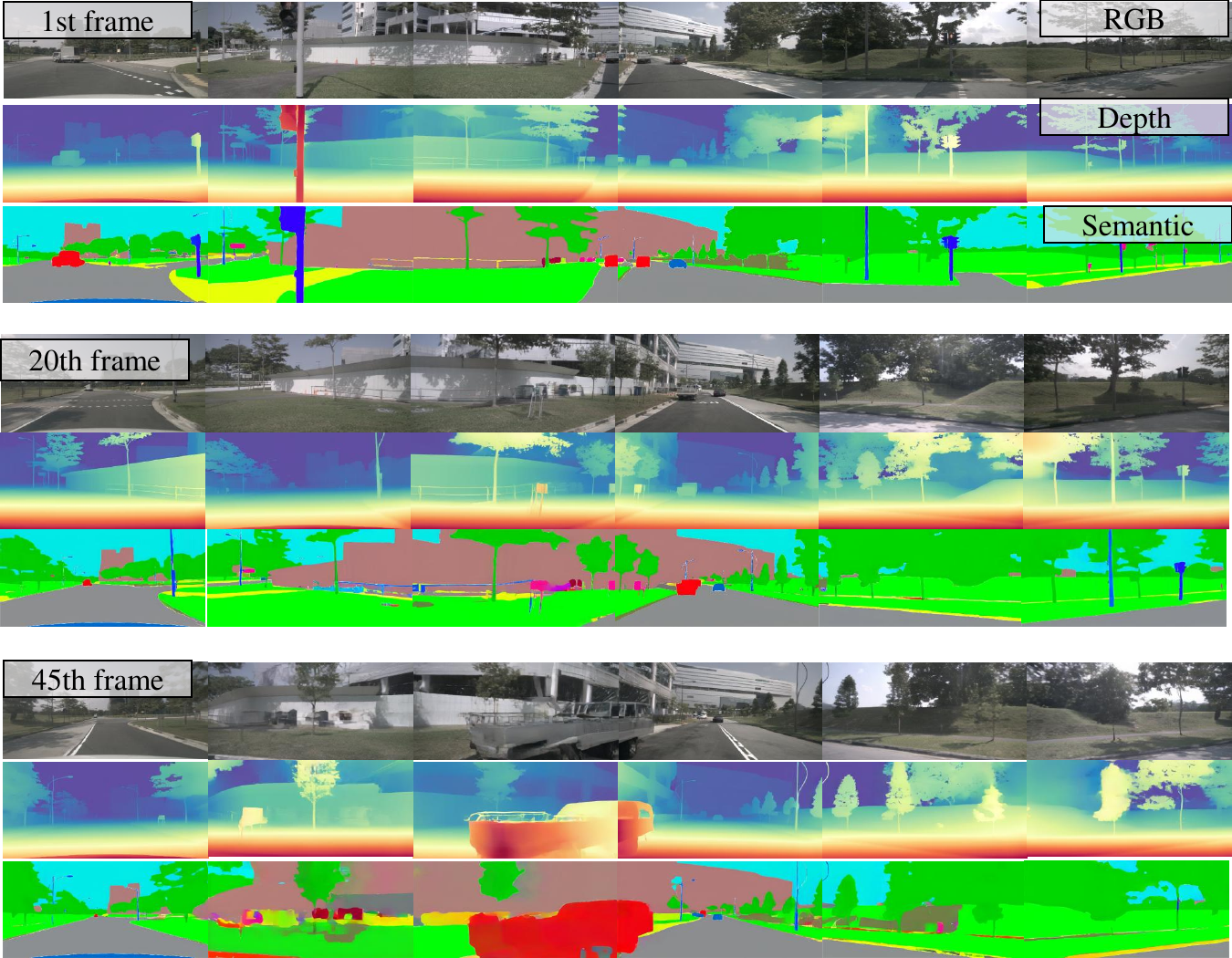}
% \vspace{-0.3cm}
\caption{Additional visualizations of our multi-view multi-modal generation results on nuScenes (arbitrarily selected frames covering different time).
}
\label{supple_fig_3}
\end{figure*}

\begin{figure*}[t]
\centering
\includegraphics[width=0.99\textwidth]{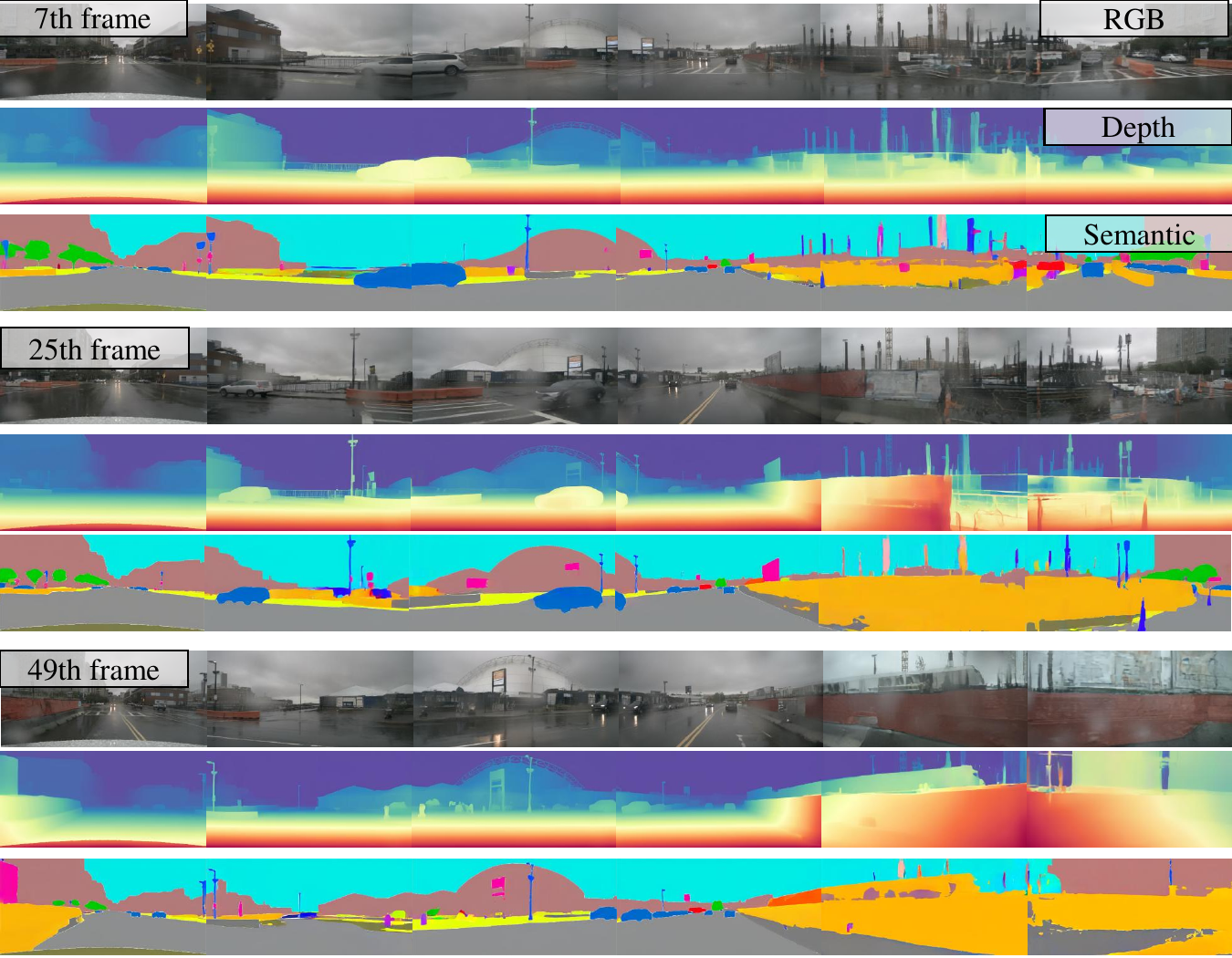}
% \vspace{-0.3cm}
\caption{Additional visualizations of our multi-view multi-modal generation results on nuScenes (arbitrarily selected frames covering different time).
}
\label{supple_fig_4}
\end{figure*}

\begin{figure*}[t]
\centering
\includegraphics[width=0.99\textwidth]{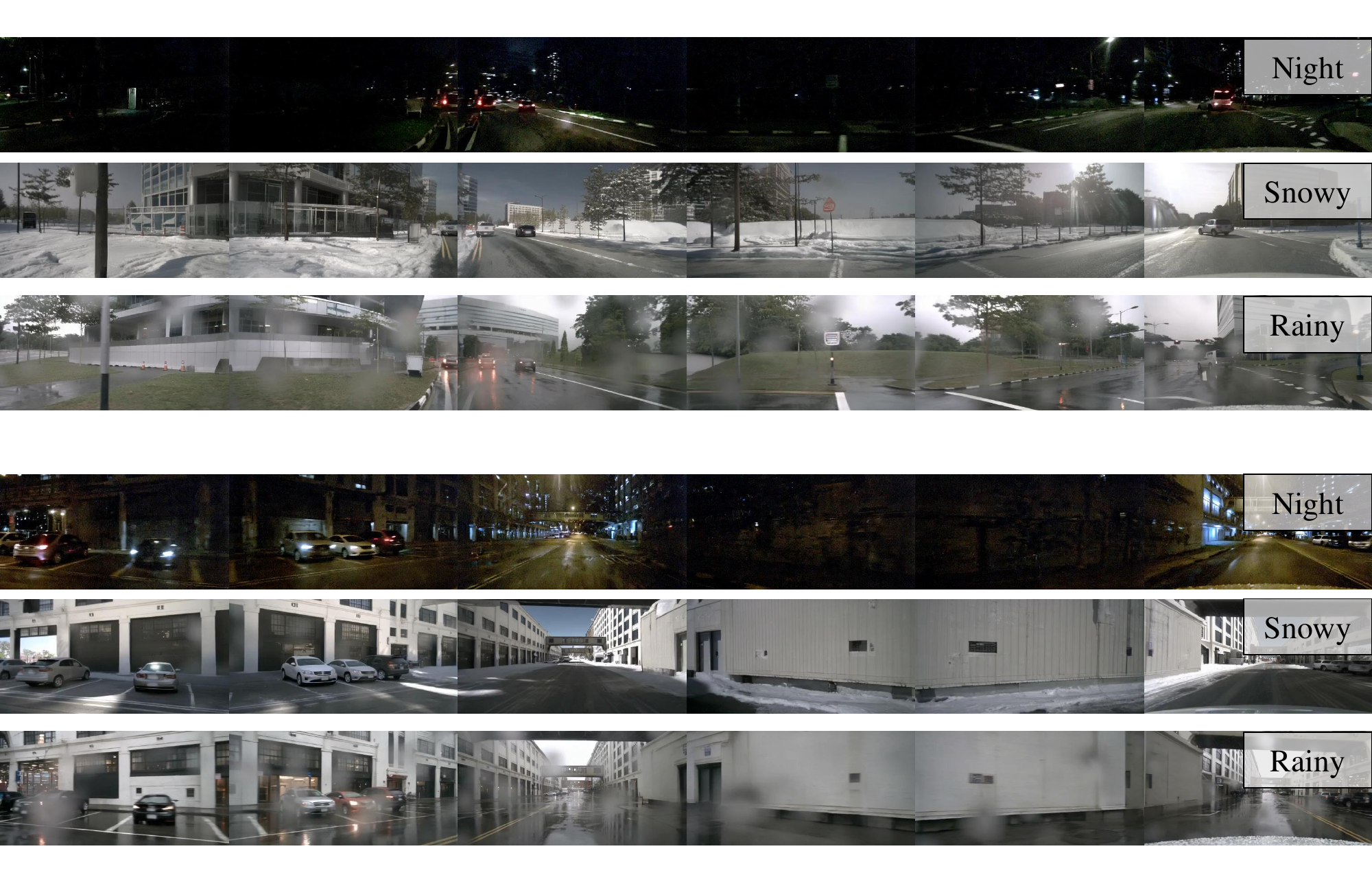}
% \vspace{-0.3cm}
\caption{Additional driving scene generation with diverse weather conditions on nuScenes.
}
\label{supple_fig_5}
\end{figure*}

\begin{figure*}[t]
\centering
\includegraphics[width=0.75\textwidth]{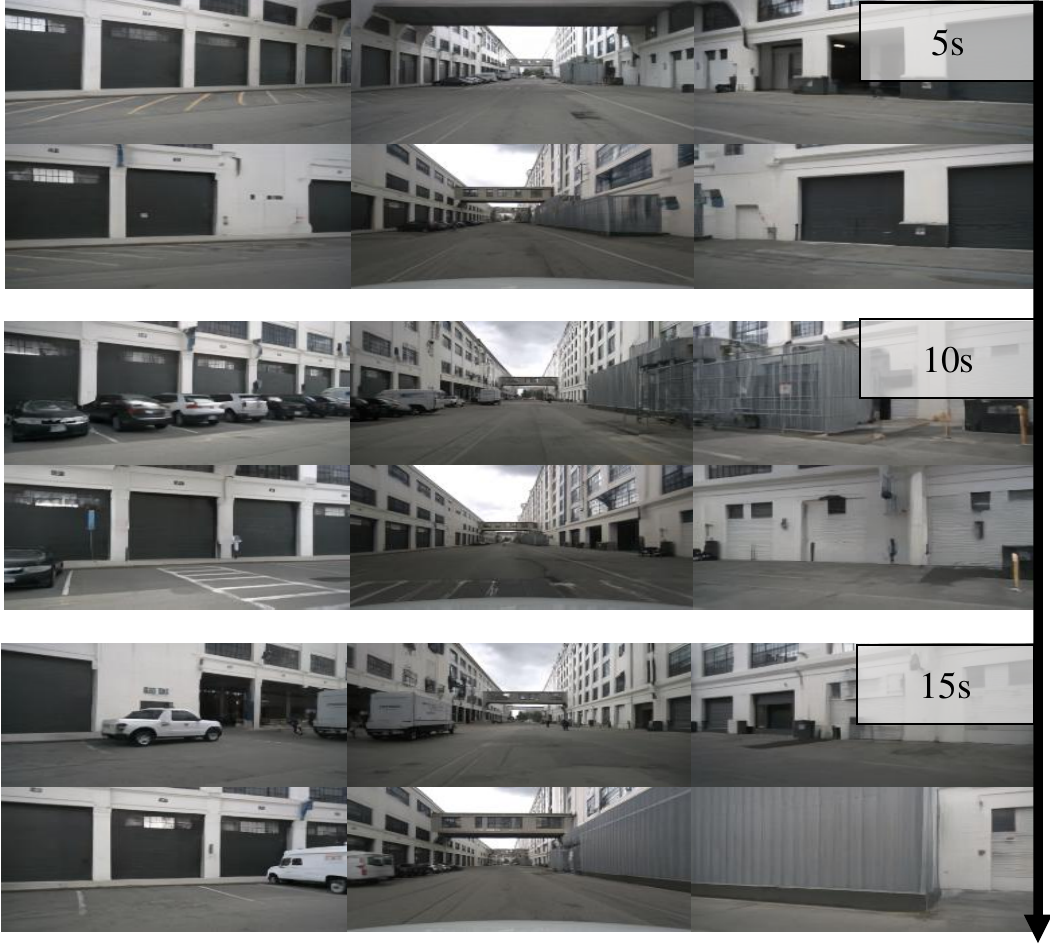}
% \vspace{-0.3cm}
\caption{Additional long video generation without reference frames on nuScenes.
}
\label{supple_fig_6}
\end{figure*}

\begin{figure*}[t]
\centering
\includegraphics[width=0.75\textwidth]{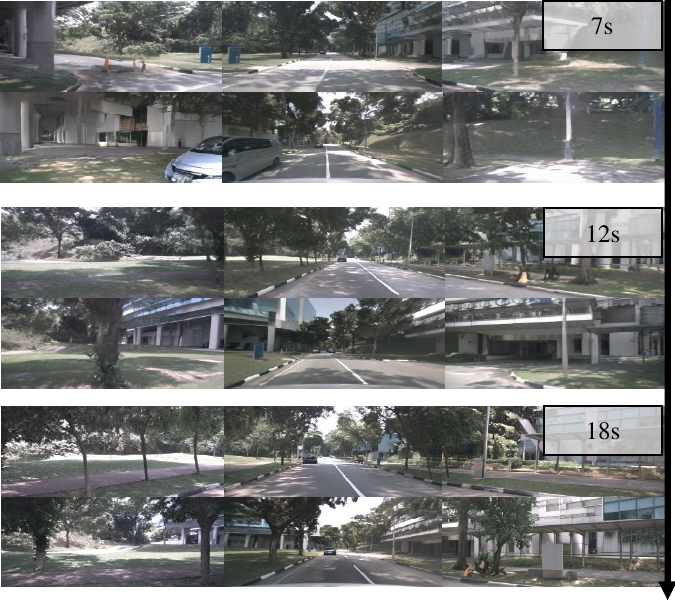}
% \vspace{-0.3cm}
\caption{Additional long video generation without reference frames on nuScenes.
}
\label{supple_fig_7}
\end{figure*}

\begin{figure*}[t]
\centering
\includegraphics[width=0.75\textwidth]{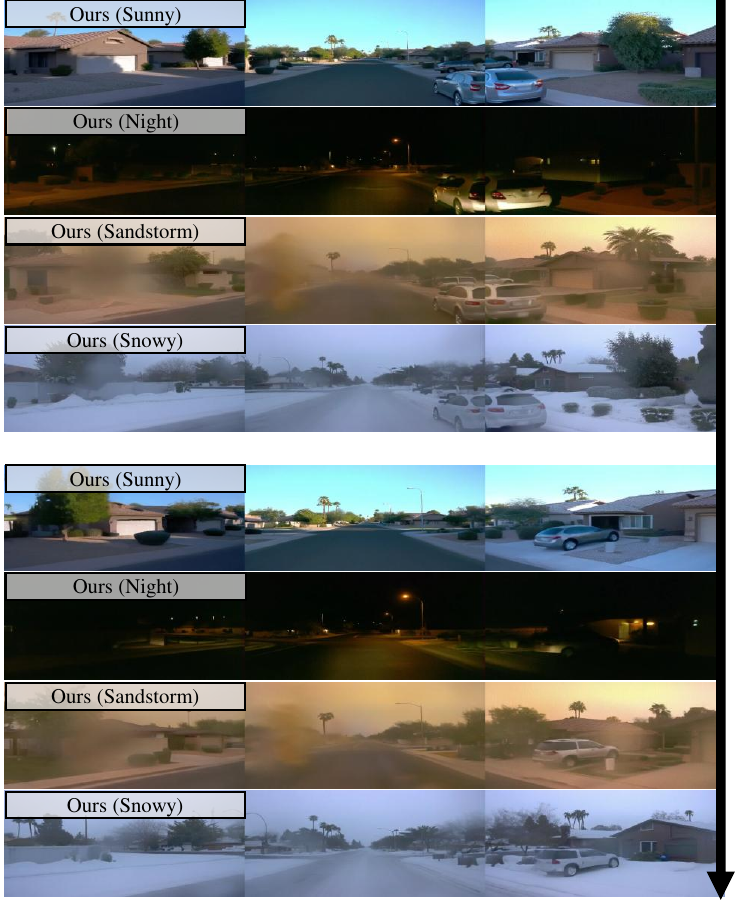}
% \vspace{-0.3cm}
\caption{Additional visualizations of experimental results on Waymo.
}
\label{supple_fig_8}
\end{figure*}

\begin{figure*}[t]
\centering
\includegraphics[width=0.69\textwidth]{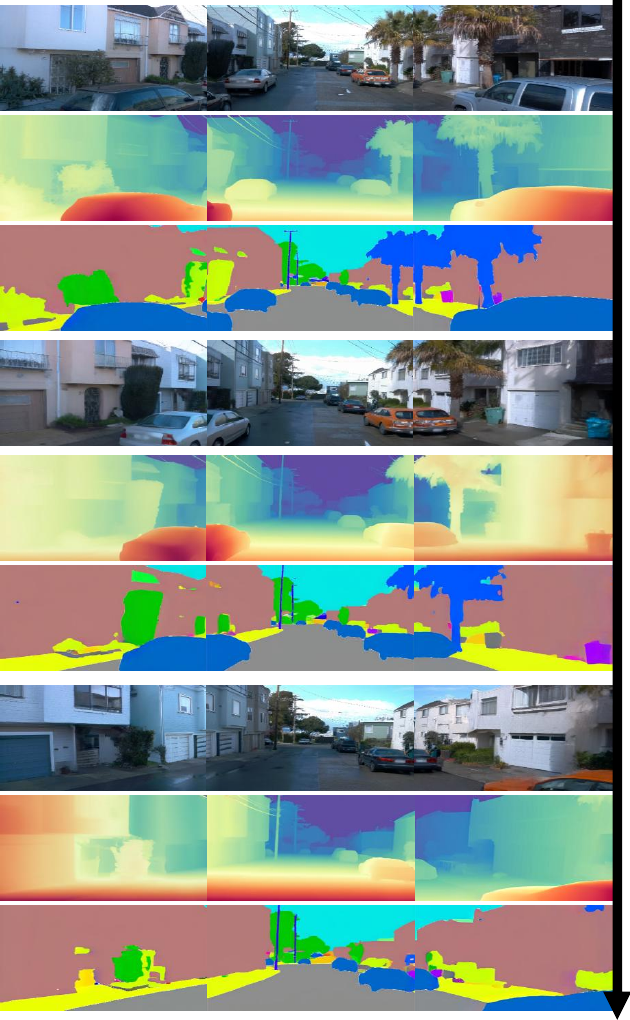}
% \vspace{-0.3cm}
\caption{Additional visualizations of our multi-view multi-modal generation results on Waymo (arbitrarily selected frames covering different time).
}
\label{supple_fig_9}
\end{figure*}

In \cref{supple_fig_1,supple_fig_2,supple_fig_3,supple_fig_4,supple_fig_5,supple_fig_6,supple_fig_7,supple_fig_8,supple_fig_9}, we show additional visualizations of our multi-view multi-modal generation results, including enlarged versions of some figures presented in the main paper.

\end{document}